\documentclass{article}
\usepackage{spconf}
\usepackage{amsmath,amssymb,amsfonts,amsthm}
\usepackage{subcaption}
\usepackage{algorithm}
\usepackage{algorithmicx}
\usepackage{algpseudocode}
\usepackage{epsfig}

\newcommand{\lb}{\left(}
\newcommand{\rb}{\right)}

\newcommand{\prox}{\textnormal{Prox}}
\newcommand{\OSCARAPO}{\textnormal{OSCAR-APO}}

\newcommand{\sign}{\textnormal{sign}}

\newcommand{\bw}{\mathbf{w}}
\newcommand{\bwt}{\widetilde{\mathbf{w}}}

\newcommand{\bx}{\mathbf{x}}
\newcommand{\by}{\mathbf{y}}

\newcommand{\bb}{\mathbf{b}}

\newcommand{\bz}{\mathbf{z}}

\newcommand{\bu}{\mathbf{u}}
\newcommand{\bg}{\mathbf{g}}

\newcommand{\bxh}{\widehat{\mathbf{x}}}
\newcommand{\bxo}{\overline{\mathbf{x}}}

\newcommand{\bnu}{\boldsymbol{\nu}}

\newcommand{\bP}{\mathbf{P}}

\newcommand{\bI}{\mathbf{I}}

\newcommand{\bA}{\mathbf{A}}

\newcommand{\cN}{\mathcal{N}}

\newcommand{\bbR}{\mathbb{R}}
\newcommand{\bbE}{\mathbb{E}}

\title{Is Ordered Weighted $\ell_1$ Regularized Regression Robust to Adversarial Perturbation? A Case Study on OSCAR}
\name{Pin-Yu Chen$^*$, Bhanukiran Vinzamuri$^*$, Sijia Liu
	 \thanks{P.-Y. Chen and B. Vinzamuri contribute equally to this work.}}
\address{IBM Research \\ \{Pin-Yu.Chen,  Bhanu.Vinzamuri, Sijia.Liu\}@ibm.com }
\begin{document}
%
\maketitle
\begin{abstract}
Many state-of-the-art machine learning models such as deep neural networks have recently shown to be vulnerable to adversarial perturbations, especially in classification tasks. Motivated by adversarial machine learning, in this paper we investigate the robustness of sparse regression models with strongly correlated covariates to adversarially designed measurement noises. Specifically, we consider the family of ordered weighted $\ell_1$ (OWL) regularized regression methods and study the case of OSCAR (octagonal shrinkage clustering algorithm for regression) in the adversarial setting. Under a norm-bounded threat model, we formulate the process of finding a maximally disruptive noise for OWL-regularized regression as an optimization problem and illustrate the steps towards finding such a noise in the case of OSCAR. Experimental results demonstrate that the regression performance of grouping strongly correlated features can be severely degraded under our adversarial setting, even when the noise budget is significantly smaller than the ground-truth signals.

\end{abstract}
\section{Introduction}
\label{sec:intro}
In recent years, adversarial machine learning has received tremendous attention as it provides new means of improving machine leaning performance and studying model robustness in the adversarial setting, such as generative adversarial networks (GANs) \cite{goodfellow2014generative} and adversarial examples \cite{goodfellow2014explaining}. In image classification tasks, well-trained machine learning models such as deep convolutional neural networks have shown to be vulnerable to adversarial examples -- human-imperceptible perturbations to natural images can be easily crafted to mislead the decision of a target image classifier \cite{szegedy2013intriguing,biggio2013evasion,biggio2017wild}, leading to new challenges on model robustness. The adversarial perturbations are often evaluated by the $\ell_1$, $\ell_2$ and $\ell_\infty$ norms \cite{kurakin2016adversarial_ICLR,carlini2017towards,chen2017ead,weng2018evaluating,su2018robustness}. Beyond image classification, other machine learning tasks such as image captioning \cite{chen2018attacking} or sequence-to-sequence text learning \cite{cheng2018seq2sick} have also shown to be vulnerable to adversarial examples. Moreover, it has been made possible to generate adversarial examples in the so-called black-box setting by simply leveraging the input-output correspondence of a target model and performing zeroth-order optimization \cite{chen2017zoo,tu2018autozoom,liu2018zeroth}.

Motivated by the recent studies in adversarial machine learning, in this paper we shift our focus to the robustness of regression models to adversarial perturbations. To the best of our knowledge, regression is a fundamental task in machine learning but little has been explored in the setting of adversarial perturbations.
 Specifically, we aim to investigate the robustness of the ordinary least-squared regression models regularized by the ordered weighted $\ell_1$ (OWL) norm \cite{bogdan2013statistical,zeng2014decreasing}.  
The OWL family of regularizers is a widely adopted method for sparse regression with strongly correlated covariates. It is worth mentioning that the octagonal shrinkage and clustering algorithm for regression \cite{bondell2008simultaneous}, which is called as OSCAR, is in fact a special case of the OWL regularizer \cite{zeng2014solving}. OSCAR is known to be more effective in identifying feature groups (i.e., strongly correlated covariates) than other feature selection methods such as LASSO \cite{tibshirani1996regression}.

In this paper, we investigate the robustness of OSCAR to adversarial perturbations by formulating the process of finding the maximally disruptive noise of the measurement model as an optimization problem. Although the recent work in \cite{figueiredo2016ordered} has established a finite-sample error upper bound on the OWL-regularized regression models associated with a norm-bounded noise level, it still remains unclear whether an adversary can disrupt the identified feature groups by intentionally manipulating the measurement error within the same noise budget. In other words, our adversarial formulation is novel in the sense that it provides a worst-case robustness analysis of OSCAR by finding a disruptive measurement error (but still within a specified noise budget) in order to deviate the detected feature groups from the ground truths. More importantly, upon verifying the lack of robustness to adversarial perturbations, our method could be incorporated to devise resilient OWL-regularized regression models via adversarial learning techniques.

Perhaps surprisingly, the experimental results show that using our proposed approach, it is possible to generate small norm-bounded perturbations to the measurement model, in order to deviate the regression results of OSCAR from the ground truths.
Consequently, our results offer new insights on the robustness analysis of OWL-regularized regression methods in the adversarial setting. 

\section{Background}
\label{sec_background}
For any real-valued vector $\bx=[x_1,x_2,\ldots,x_p] \in \bbR^{p}$, let $|\bx|$ denote the vector of its element-wise absolute value  and let $|\bx|^\downarrow$ denote the element-permuted vector of $|\bx|$ such that  $|x|^\downarrow_1 \geq |x|^\downarrow_2 \geq \ldots \geq |x|^\downarrow_p$, where $|x|^\downarrow_i$ is the $i$-th largest component of $\bx$. Given a vector $\bw \in \bbR^p_+$ such that $w_1 \geq w_2 \geq \ldots \geq 0$ and $w_1>0$, the OWL norm \cite{bogdan2013statistical,zeng2014decreasing} is defined as 
\begin{align}
\label{eqn_OWL}
\Omega_{\bw}(\bx)=\sum_{i=1}^p w_i |x|^\downarrow_i.
\end{align}
The OSCAR regularizer \cite{bondell2008simultaneous} is a special case of the OWL norm in \eqref{eqn_OWL} when $w_i=\lambda_1 + \lambda_2 (p-i)$, where $\lambda_1, \lambda_2 \geq 0$.

We consider the OWL-regularized linear regression problem taking the following form:
\begin{align}
\label{eqn_OWL_regression}
\textnormal{Minimize~}_{\bx \in \bbR^p} \|\by-\bA \bx \|_2^2 + \lambda \Omega_{\bw}(\bx),
\end{align}
where $\by \in \bbR^n$ is the vector of $n$ noisy measurements, $\bA \in \bbR^{n \times p}$ is the design matrix, and $\lambda \geq 0$ is the regularization parameter of the OWL norm.

The seminal work in \cite{figueiredo2016ordered} establishes a finite-sample error bound on the OWL-regularized regression method under the measurement model
\begin{align}
\label{eqn_measurement_model}
\by=\bA \bx^* + \bnu,
\end{align}
where $\bx^* \in \bbR^p$ is an $s$-sparse vector (i.e., the signal) and $\bnu \in \bbR^n$ is the measurement error (i.e., the noise). Let $\bxo^*$ denote the vector with identical coefficients corresponding to identical columns in $\bA$, and assume the entries in each column of $\bA$ are i.i.d. $\cN(0,1)$ (standard Gaussian random variables) but different columns could be strongly correlated.
Consider the $\ell_1$-norm bounded measurement error constraint $\frac{\| \bnu\|_1}{n} \leq \epsilon $, then the solution $\bxh$ to \eqref{eqn_OWL_regression} is guaranteed to satisfy the finite-sample error upper bound \cite{figueiredo2016ordered}:
\begin{align}
\label{eqn_upper_bound}
\bbE \|\bxh - \bxo^* \|_2 \leq C_1 \lb C_2 \|\bx^*\|_2  \sqrt{\frac{s \log p}{n}} + \epsilon \rb, 
\end{align}
where $\|\cdot\|_2$ denotes the $\ell_2$ norm, the expectation $\bbE$ is taken over the random design matrix $\bA$, and $C_1$ and $C_2$ are some positive constants that we omitted for brevity (see Theorem 1.1 in \cite{figueiredo2016ordered}). Note that in this finite-sample analysis no distributional assumptions are imposed on the measurement error $\bnu$ other than its bounded $\ell_1$ norm.
Similar error bound can be obtained when the rows of $\bA$ are i.i.d. Gaussian random vectors \cite{figueiredo2016ordered}.

In general, the solution $\bxh$ to \eqref{eqn_OWL_regression} can be efficiently obtained by leveraging the proximal operator $\prox_{\Omega_{\bw}}(\cdot)$ of OWL regularization. As illustrated in \cite{zeng2014ordered}, one can use the fast iterative shrinkage-thresholding algorithm (FISTA) \cite{beck2009fast} to obtain $\bxh$, which includes iterating the following optimization steps: 
\begin{enumerate}
	\item OWL proximal gradient descent -- \\ $\bx^{(k)}=\prox_{\lambda \Omega_{\bw}} \lb \bu^{(k)} - \bA^T (\bA \bu^{(k)} - \by) / \alpha_k \rb$
	\item Momentum -- $\bu^{(k+1)}=\bx^{(k)} + \beta_k \lb \bx^{(k)} - \bx^{(k-1)} \rb $
	\item Update $\alpha_k$ and $\beta_k$ if not converged
\end{enumerate}
The index $k$ denotes the FISTA iteration, and $\alpha_k$ and $\beta_k$ denote the inverse of the step size and the momentum coefficient, respectively. The notation $\cdot^T$ denotes matrix transpose.

Specifically, when the OWL regularizer reduces to OSCAR, its approximate proximity operator (APO) has a closed-form expression given by \cite{zeng2014solving}
\begin{align}
\label{eqn_APO}
\prox_{\OSCARAPO}(\bz)=\sign(\bz)  \odot \max \{ |\bz| - \bwt,0 \}, 
\end{align}
where $\sign(\cdot)$ is the vector of entry-wise sign function ($+1$ or $-1$),  $\odot$ denotes entry-wise product, $\max\{\bz,0\}$ denotes entry-wise maximum value between $z_i$ and $0$, and $\widetilde{\bw} = \bP(\bz)^T \bw$, where $\bP(\bz)$ is a permutation matrix associated with a given vector $\bz \in \bbR^p$ satisfying $\bP(\bz) |\bz|= |\bz|^\downarrow$, which can be obtained by taking $|\bz|$ and sorting its entries in descending order. In addition, for OSCAR the vector $\bw$ in OWL has the relation $w_i=\lambda_1 + \lambda_2 (p-i)$, where $\lambda_1, \lambda_2 \geq 0$.

\section{Main Results}
Although a finite-sample analysis for OWL-regularized regression has been established under the $\ell_1$-norm constrained measurement error $\epsilon$ in \cite{figueiredo2016ordered}, motivated by the recent advances in adversarial machine learning, we are interested in investigating its robustness to  adversarially designed noises satisfying the same error budget $\epsilon$. In other words, given an $\ell_1$-norm bounded threat model $\frac{\|\bnu\|_1}{n} \leq \epsilon$ and a design matrix $\bA$, we aim to find an optimal noise $\bnu^*$ that could maximally degrade the performance of OWL-regularized regression in terms of the detected  feature groups. In this paper,
we particularly focus on the case of OSCAR with APO as its solver.

Under the same measurement model as in \eqref{eqn_measurement_model}, we formulate the problem of finding a norm-bounded noise $\bnu^*$ that could maximally deviate the OWL-regularized regression $\bxh$ from the ground-truth feature cluster membership vector $\bxo^*$ by solving
\begin{align}
\label{eqn_Adv_OWL}
&\textnormal{Maximize}_{\bnu \in \bbR^n}~  \| \bxh(\bnu) - \bxo^* \|_2^2 \\
&\textnormal{subject~to~} \|\bnu\|_1 / n \leq \epsilon, \\ &\bxh(\bnu)=\arg \min_{\bx \in \bbR^p} \| \by - \bA \bx \|_2^2 + \lambda \Omega_{\bw}(\bx).
\label{eqn_Adv_OWL_2}
\end{align}
Essentially, our adversarial formulation studies the robustness of OWL-regularized regression in the worst-case scenario by exploring the space of constrained measurement noise to maximize the feature group identification loss in \eqref{eqn_Adv_OWL}.
In our setting, we assume the adversary has access to the ground-truth vector $\bx^*$ so that based on \eqref{eqn_measurement_model}, \eqref{eqn_Adv_OWL_2} can be written as 
\begin{align}
\label{eqn_Adv_OWL_ground}
\bxh(\bnu)=\arg \min_{\bx \in \bbR^p} \| \bA (\bx^*-\bx) + \bnu \|_2^2 + \lambda \Omega_{\bw}(\bx).
\end{align}

Next we specify how to solve the adversarial regression formulation in \eqref{eqn_Adv_OWL} to \eqref{eqn_Adv_OWL_2}
in the case of OSCAR with APO as its solver. Let $\bu^*$ and $\alpha^*$ be the final iterates of the momentum and step size terms in FISTA as described in Section \ref{sec_background}. Given the measurement model \eqref{eqn_measurement_model} under OSCAR-APO, \eqref{eqn_Adv_OWL_ground} becomes
\begin{align}
\label{eqn_OSCAR_APO}
\bxh(\bnu)&=\prox_{\OSCARAPO} \lb \bu^* - \bA^T (\bA \bu^*- \by)/ \alpha^* \rb \nonumber \\
&=\prox_{\OSCARAPO} \lb \bu^* - \bA^T (\bA \bu^*- \bA \bx^* - \bnu)/ \alpha^* \rb,
\end{align}
where $\prox_{\OSCARAPO}(\cdot)$ is defined in \eqref{eqn_APO}.
With the method of Lagrange multipliers, we are interested in solving the following alternative optimization problem
\begin{align}
\label{eqn_Adv_OWL_Lagrange}
&\textnormal{Minimize}_{\bnu \in \bbR^n}   -\| \bxh(\bnu) - \bxo^* \|_2^2 + \frac{\gamma}{n} \|\bnu\|_1,
\end{align}
where $\gamma >0$ is a tunable regularization coefficient such that the solution $\bnu^*$ to \eqref{eqn_Adv_OWL_Lagrange} will satisfy the norm constraint $\|\bnu^*\|_1/n \leq \epsilon$.

We note that the formulation in \eqref{eqn_Adv_OWL_Lagrange} falls into the category of LASSO problem \cite{tibshirani1996regression} and can be efficiently solved by using optimization methods such as iterative shrinkage-thresholding algorithm (ISTA). Specifically, let $f(\bnu)= -\| \bxh(\bnu) - \bxo^* \|_2^2$. Then  $\bnu^*$ (possibly a local optimum) can be obtained by iteratively solving
\begin{align}
\bnu^{(k+1)}=S_{\gamma/n} \lb \bnu^{(k)} - \eta_k \nabla f(\bnu^{(k)}) \rb,
\end{align}
where $\bnu^{(k)}$ denotes the $k$-th iterate, $\eta_k$ is the step size, $\nabla f$ denotes the gradient of $f$\footnote{We use the subgradient of $f$ at points where $f$ is not differentiable.}, and $S_{\rho}(\bz): \bbR^n \mapsto \bbR^n$ is an entry-wise function defined as 
\begin{align}
S_{\rho}(z_i)= 
\left\{
\begin{array}{ll}
z_i - \rho, & \text{~if~} z_i > \rho ; \\
0, & \text{~if~} |z_i| \leq \rho ; \\
-z_i + \rho, & \text{~if~} z_i < -\rho.
\end{array}
\right.
\end{align}

In what follows, we explicitly derive the gradient  $\nabla f(\bnu)=[\frac{\partial f}{\partial \nu_1},\frac{\partial f}{\partial \nu_2},\ldots,\frac{\partial f}{\partial \nu_n}]^T$  of $f(\bnu)$ with respect to $\bnu$ for OSCAR-APO. For clarity, the index $i \in \{1,\ldots,n\}$ specifies the measurement instance, and  the index $j \in \{1,\ldots,p\}$ specifies the covariate instance. To simplify the notation, let $\bb(\bnu)=\bu^* - \bA^T (\bA \bu^*- \bA \bx^* - \bnu)/ \alpha^*$ such that $\widehat{\bx}(\bnu)=\prox_{\OSCARAPO}(\bb)=\textnormal{sign}(\bb) \odot \max \{ |\bb| - \widetilde{\bw},0 \} $ based on \eqref{eqn_APO}, where $\widetilde{\bw}=\bP(\bb)^T \bw$. Rewriting $f(\bnu)=-\sum_{j=1}^p ([\widehat{\bx}(\bnu)]_j - \overline{x}^*_j)^2=-\sum_{j=1}^p  [ \sign(b_j) \cdot \max\{|b_j| - \widetilde{w}_j,0\} - \overline{x}^*_j]^2$ and using chain rule, we can obtain
\begin{align}
\label{eqn_partial_der}
\frac{\partial f}{\partial \nu_i} &= \frac{\partial f}{\partial \bb}^T  \frac{\partial \bb}{\partial \nu_i} \nonumber \\
& =  \sum_{j=1}^p \frac{\partial f}{\partial b_j} \cdot  \frac{\partial b_j}{\partial \nu_i} \nonumber \\
&= \sum_{j:|b_j| > \widetilde{w}_j} -2  \lb b_j - \sign(b_j) \widetilde{w}_j - \overline{x}^*_j \rb \cdot \frac{A_{ij}}{\alpha^*} \nonumber \\
&= \sum_{j=1}^p -2  \lb b_j - \sign(b_j) \widetilde{w}_j - \overline{x}^*_j \rb \cdot \frac{A_{ij}}{\alpha^*} \cdot I_{\{|b_i| > \widetilde{w}_j\}}, 
\end{align}
where $I_E$ is an indicator function such that  $I_E=1$ if event $E$ is true; otherwise $I_E=0$.

\begin{algorithm}[t]
	\caption{Adversarial perturbation for OSCAR-APO}
	\label{algo_OSCAR_APO}
	\begin{algorithmic}
		\State 	\textbf{Input:} $\bA$, $\bx^*$, $\bxo^*$, $\bw$, $\bu^*$, $\alpha^*$, $\epsilon$, $\{\eta_k\}$
		\State \textbf{Output: $\bnu^*$}
		\State \textbf{Initialization:} $k=0$, $\gamma=\gamma_0$, $\bg \sim \cN(0,\bI_n)$ and 
		$\bnu^{(0)} = \epsilon \cdot \bg/\|\bg\|_1 $
		\While{not converged}
		\State 1. $\bb=\bu^* - \bA^T \lb \bA \bu^*- \bA \bx^* - \bnu^{(k)} \rb / \alpha^*$
		\State 2. Find the permutation $\bP(\bb)$ s.t. $\bP(\bb)|\bb|=|\bb|^\downarrow$
		\State 3. $\bwt =  \bP(\bb)^T \bw$
		\State 4. $\frac{\partial f}{\partial \nu_i}= \sum_{j:|b_j| > \widetilde{w}_j} -2  \lb b_j - \sign(b_j) \widetilde{w}_j - \overline{x}^*_j \rb \cdot \frac{A_{ij}}{\alpha^*}$  
		\State ~~~~~for all $i \in \{1,\ldots,n\}$
		\State 5. $\bnu^{(k+1)}=S_{\gamma/n} \lb \bnu^{(k)} - \eta_k \nabla f(\bnu^{(k)}) \rb$
		\State 6. $k \leftarrow k+1$
		\EndWhile
		\State $\bnu^* \leftarrow \bnu^{(k)}$
		\If{ $\|\bnu^*\|_1 / n > \epsilon$}
		\State Reinitialization with a larger $\gamma$ and redo the while loop
		\EndIf
	\end{algorithmic}
\end{algorithm}

\begin{figure*}
\centering
\includegraphics[scale=0.9]{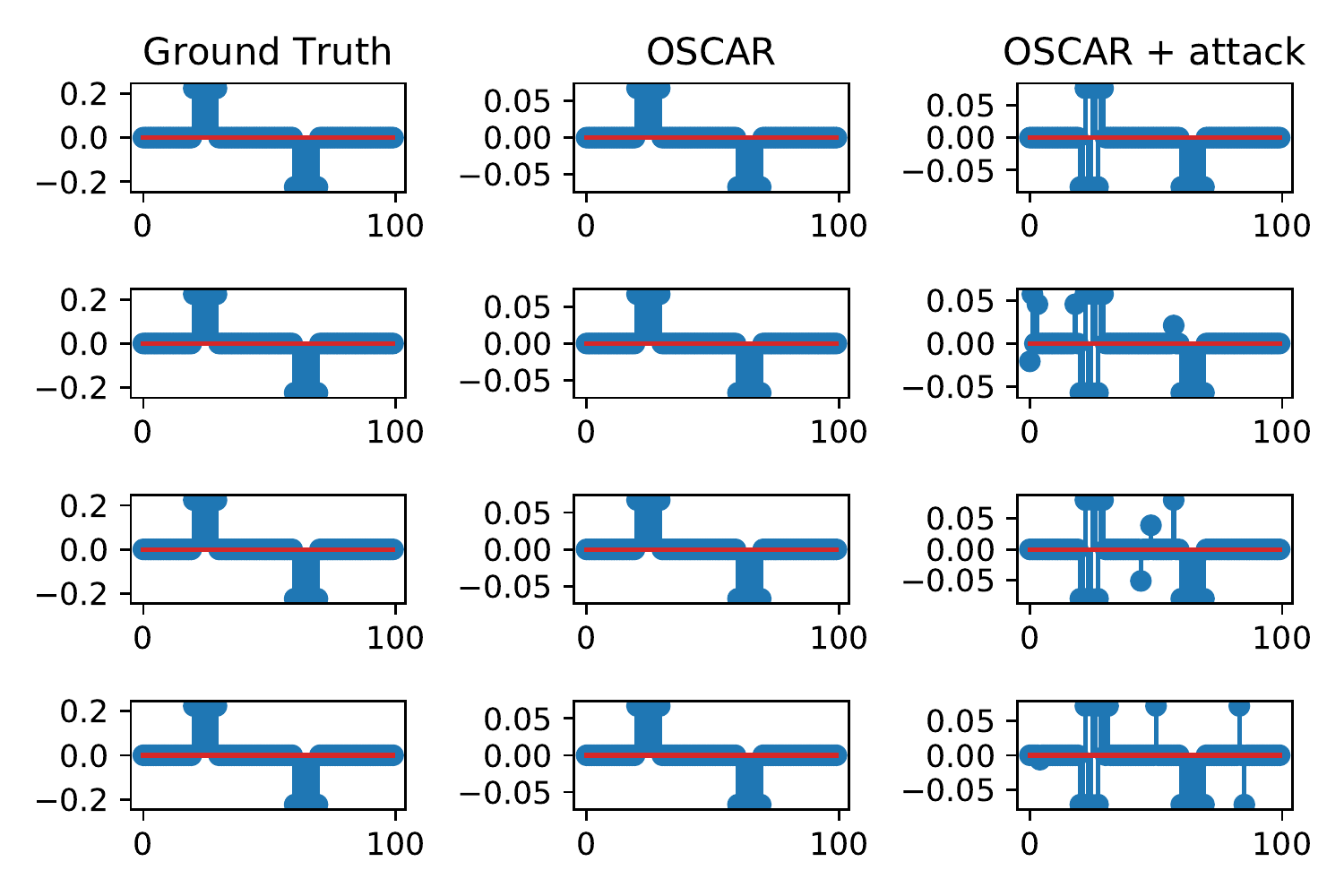}
\vspace{-6mm}
\caption{Assessing effect of our proposed adversarial attack on OSCAR (rightmost column) by varying the attack strength $\epsilon$ with  $\epsilon=0.05$ (top row), $\epsilon=0.1$ (second row), $\epsilon=0.2$ (third row) and  $\epsilon=0.3$ (fourth row). In each column, the ground truth refers to $\bx^*$, OSCAR refers to the regression results without noise, and OSCAR + Attack refers to the regression results against our designed adversarial noises. The feature groups can be adversarially misaligned even for small $\epsilon$.}
\label{fig:oscarattack}
\vspace{-2mm}
\end{figure*}

The detailed derivations are as follows. To obtain $\frac{\partial f}{\partial b_j}$, we divide the analysis into three cases based on the value of $b_j$:
\begin{itemize}
	\item Case I -- If $|b_j| \leq \widetilde{w}_j $, then $[\widehat{\bx}(\bnu)]_j = 0$ and hence $\frac{\partial f}{\partial b_j}=0$.
	\item Case II -- If $b_j > \widetilde{w}_j $, then $[\widehat{\bx}(\bnu)]_j=\textnormal{sign}(b_j) \cdot (b_j - \widetilde{w}_j)=b_j - \widetilde{w}_j$ since $\textnormal{sign}(b_j)=1$. Therefore, $\frac{\partial f}{\partial b_j}=-2  (b_j - \widetilde{w}_j - \overline{x}^*_j).$ We note that technically, $\bwt$ is also a function of $\bb$ and hence a function of $\bnu$. As a result, one needs further chain rule factorization $\frac{\partial f}{\partial b_j} = \frac{\partial f}{\partial~b_j - \widetilde{w}_j}  \cdot \frac{\partial~b_j - \widetilde{w}_j}{\partial b_j}$. Here we implicitly use the fact that $\frac{\partial \widetilde{w}_j}{\partial b_j}=\frac{\partial [\bP(\bb)^T \bw]_j}{\partial b_j}=0$ since no matter how we permute the entries of $\bw$, it is still a constant vector.
	\item Case III --	If $-b_j> \widetilde{w}_j $, then $[\widehat{\bx}(\bnu)]_j= \textnormal{sign}(b_j) \cdot (-b_j - \widetilde{w}_j)=b_j + \widetilde{w}_j$. Similar to the analysis of Case II, we obtain
	$\frac{\partial f}{\partial b_j}=-2  (b_j + \widetilde{w}_j - \overline{x}^*_j) $.
\end{itemize}
Summarizing these cases, we have a simplified expression $\frac{\partial f}{\partial b_j}=-2  (b_j - \sign(b_j) \widetilde{w}_j - \overline{x}^*_j) $ if $|b_j|>\widetilde{w}_j$ and  $\frac{\partial f}{\partial b_j}=0$ otherwise. Next, to obtain $\frac{\partial b_j}{\partial \nu_i}$, based on the definition of $\bb$ we have $b_j = \textnormal{constant} + \sum_{k=1}^n A^T_{jk} \nu_k/\alpha^*=\sum_{k=1}^n A_{kj} \nu_k/\alpha^*$. Therefore, $\frac{\partial b_j}{\partial \nu_i}=A_{ij}/\alpha^*$. Finally, combining the analysis of $\frac{\partial f}{\partial b_j}$ and 
$\frac{\partial b_j}{\partial \nu_i}$, we obtain the results in \eqref{eqn_partial_der}. 

The pipeline of crafting a norm-bounded adversarial perturbation $\bnu^*$ for OSCAR-APO is describe in Algorithm \ref{algo_OSCAR_APO}. We also note that beyond OSCAR-APO, it is possible to craft an adversarial noise $\bnu^*$ for generic OWL-regularized regression methods by treating \eqref{eqn_Adv_OWL_2} as a black-box function, which will be considered in our future work.

\section{Performance Evaluation}
In this experiment, we generated a synthetic dataset of $p=100$ features and $n=50$ instances with a pre-defined grouping structure among the features. The features (entries in $\bA$) were generated using a standard Gaussian distribution and we modeled the response variable using \eqref{eqn_measurement_model}. Given a norm-bounded noise strength $\epsilon$, we call the adversarial perturbation found using our proposed approach (Algorithm \ref{algo_OSCAR_APO}) an ``attack'' on the considered regression method. For the attack parameters, we set $\gamma_0=0.1$ and $\eta_k=10^{-3}$.
In Figure~\ref{fig:oscarattack}, the x-axis represents the feature index and the y-axis represents the coefficient values. The left column represents the ground-truth $\bx^*$ with two defined feature groups. The middle column shows the feature grouping obtained after running OSCAR algorithm in the noiseless setting. The right column shows how the grouping is adversely affected after our attack. 
We varied the noise budget $\epsilon$ from 0.05 to 0.3 to assess the effect of the attack. One can observe that although some grouped features are retained up to a certain degree, the true effect of the attack can be seen on the features which are misaligned from their original feature groups, even for relatively small $\epsilon$.

\section{Conclusion and Future Work}
To study the robustness of OWL-regularized regression methods in the adversarial setting, this paper proposes a novel formulation for finding norm-bounded adversarial perturbations in the measurement model and illustrates the pipeline of adversarial noise generation in the case of OSCAR with APO as its solver. In the adversarial setting, the experimental results show that our proposed approach can effectively craft adversarial noises that severely degrade the regression performance in identifying ground-truth grouped features, even in the regime of small noise budgets. Our results indicate the potential risk of lacking robustness to adversarial noises in the tested regression method. One possible extension of our approach is to devise adversary-resilient regression methods. Our future work also includes developing a generic framework for generating adversarial noises for the entire family of OWL-regularized regression methods.

\vfill\pagebreak

\clearpage


\bibliography{IEEEabrv,adversarial_learning}

\begin{thebibliography}{10}
\providecommand{\url}[1]{#1}
\csname url@samestyle\endcsname
\providecommand{\newblock}{\relax}
\providecommand{\bibinfo}[2]{#2}
\providecommand{\BIBentrySTDinterwordspacing}{\spaceskip=0pt\relax}
\providecommand{\BIBentryALTinterwordstretchfactor}{4}
\providecommand{\BIBentryALTinterwordspacing}{\spaceskip=\fontdimen2\font plus
\BIBentryALTinterwordstretchfactor\fontdimen3\font minus
  \fontdimen4\font\relax}
\providecommand{\BIBforeignlanguage}[2]{{%
\expandafter\ifx\csname l@#1\endcsname\relax
\typeout{** WARNING: IEEEtran.bst: No hyphenation pattern has been}%
\typeout{** loaded for the language `#1'. Using the pattern for}%
\typeout{** the default language instead.}%
\else
\language=\csname l@#1\endcsname
\fi
#2}}
\providecommand{\BIBdecl}{\relax}
\BIBdecl

\bibitem{goodfellow2014generative}
I.~Goodfellow, J.~Pouget-Abadie, M.~Mirza, B.~Xu, D.~Warde-Farley, S.~Ozair,
  A.~Courville, and Y.~Bengio, ``Generative adversarial nets,'' in
  \emph{Advances in neural information processing systems}, 2014, pp.
  2672--2680.

\bibitem{goodfellow2014explaining}
I.~J. Goodfellow, J.~Shlens, and C.~Szegedy, ``Explaining and harnessing
  adversarial examples,'' \emph{ICLR, arXiv preprint arXiv:1412.6572}, 2015.

\bibitem{szegedy2013intriguing}
C.~Szegedy, W.~Zaremba, I.~Sutskever, J.~Bruna, D.~Erhan, I.~Goodfellow, and
  R.~Fergus, ``Intriguing properties of neural networks,'' \emph{ICLR, arXiv
  preprint arXiv:1312.6199}, 2014.

\bibitem{biggio2013evasion}
B.~Biggio, I.~Corona, D.~Maiorca, B.~Nelson, N.~{\v{S}}rndi{\'c}, P.~Laskov,
  G.~Giacinto, and F.~Roli, ``Evasion attacks against machine learning at test
  time,'' in \emph{Joint European conference on machine learning and knowledge
  discovery in databases}, 2013, pp. 387--402.

\bibitem{biggio2017wild}
B.~Biggio and F.~Roli, ``Wild patterns: Ten years after the rise of adversarial
  machine learning,'' \emph{arXiv preprint arXiv:1712.03141}, 2017.

\bibitem{kurakin2016adversarial_ICLR}
A.~Kurakin, I.~Goodfellow, and S.~Bengio, ``Adversarial machine learning at
  scale,'' \emph{ICLR, arXiv preprint arXiv:1611.01236}, 2017.

\bibitem{carlini2017towards}
N.~Carlini and D.~Wagner, ``Towards evaluating the robustness of neural
  networks,'' in \emph{IEEE Symposium on Security and Privacy}, 2017, pp.
  39--57.

\bibitem{chen2017ead}
P.-Y. Chen, Y.~Sharma, H.~Zhang, J.~Yi, and C.-J. Hsieh, ``{EAD}: elastic-net
  attacks to deep neural networks via adversarial examples,'' \emph{AAAI, arXiv
  preprint arXiv:1709.04114}, 2018.

\bibitem{weng2018evaluating}
T.-W. Weng, H.~Zhang, P.-Y. Chen, J.~Yi, D.~Su, Y.~Gao, C.-J. Hsieh, and
  L.~Daniel, ``Evaluating the robustness of neural networks: An extreme value
  theory approach,'' \emph{ICLR, arXiv preprint arXiv:1801.10578}, 2018.

\bibitem{su2018robustness}
D.~Su, H.~Zhang, H.~Chen, J.~Yi, P.-Y. Chen, and Y.~Gao, ``Is robustness the
  cost of accuracy?--a comprehensive study on the robustness of 18 deep image
  classification models,'' \emph{ECCV, arXiv preprint arXiv:1808.01688}, 2018.

\bibitem{chen2018attacking}
H.~Chen, H.~Zhang, P.-Y. Chen, J.~Yi, and C.-J. Hsieh, ``Attacking visual
  language grounding with adversarial examples: A case study on neural image
  captioning,'' in \emph{Proceedings of the 56th Annual Meeting of the
  Association for Computational Linguistics}, vol.~1, 2018, pp. 2587--2597.

\bibitem{cheng2018seq2sick}
M.~Cheng, J.~Yi, H.~Zhang, P.-Y. Chen, and C.-J. Hsieh, ``Seq2sick: Evaluating
  the robustness of sequence-to-sequence models with adversarial examples,''
  \emph{arXiv preprint arXiv:1803.01128}, 2018.

\bibitem{chen2017zoo}
P.-Y. Chen, H.~Zhang, Y.~Sharma, J.~Yi, and C.-J. Hsieh, ``{ZOO}: Zeroth order
  optimization based black-box attacks to deep neural networks without training
  substitute models,'' in \emph{ACM Workshop on Artificial Intelligence and
  Security}, 2017, pp. 15--26.

\bibitem{tu2018autozoom}
C.-C. Tu, P.~Ting, P.-Y. Chen, S.~Liu, H.~Zhang, J.~Yi, C.-J. Hsieh, and S.-M.
  Cheng, ``Autozoom: Autoencoder-based zeroth order optimization method for
  attacking black-box neural networks,'' \emph{arXiv preprint
  arXiv:1805.11770}, 2018.

\bibitem{liu2018zeroth}
S.~Liu, B.~Kailkhura, P.-Y. Chen, P.~Ting, S.~Chang, and L.~Amini,
  ``Zeroth-order stochastic variance reduction for nonconvex optimization,''
  \emph{arXiv preprint arXiv:1805.10367}, 2018.

\bibitem{bogdan2013statistical}
M.~Bogdan, E.~van~den Berg, W.~Su, and E.~J. Cand{\`e}s, \emph{Statistical
  estimation and testing via the ordered ${\ell_1}$ norm}.\hskip 1em plus 0.5em
  minus 0.4em\relax STANFORD University, 2013.

\bibitem{zeng2014decreasing}
X.~Zeng and M.~A. Figueiredo, ``Decreasing weighted sorted ${\ell_1}$
  regularization,'' \emph{IEEE Signal Processing Letters}, vol.~21, no.~10, pp.
  1240--1244, 2014.

\bibitem{bondell2008simultaneous}
H.~D. Bondell and B.~J. Reich, ``Simultaneous regression shrinkage, variable
  selection, and supervised clustering of predictors with oscar,''
  \emph{Biometrics}, vol.~64, no.~1, pp. 115--123, 2008.

\bibitem{zeng2014solving}
X.~Zeng and M.~A. Figueiredo, ``Solving oscar regularization problems by fast
  approximate proximal splitting algorithms,'' \emph{Digital Signal
  Processing}, vol.~31, pp. 124--135, 2014.

\bibitem{tibshirani1996regression}
R.~Tibshirani, ``Regression shrinkage and selection via the lasso,''
  \emph{Journal of the Royal Statistical Society. Series B (Methodological)},
  pp. 267--288, 1996.

\bibitem{figueiredo2016ordered}
M.~Figueiredo and R.~Nowak, ``Ordered weighted ${\ell_1}$ regularized
  regression with strongly correlated covariates: Theoretical aspects,'' in
  \emph{Artificial Intelligence and Statistics}, 2016, pp. 930--938.

\bibitem{zeng2014ordered}
X.~Zeng and M.~A. Figueiredo, ``The ordered weighted ${\ell_1}$ norm: Atomic
  formulation, projections, and algorithms,'' \emph{arXiv preprint
  arXiv:1409.4271}, 2014.

\bibitem{beck2009fast}
A.~Beck and M.~Teboulle, ``A fast iterative shrinkage-thresholding algorithm
  for linear inverse problems,'' \emph{SIAM journal on imaging sciences},
  vol.~2, no.~1, pp. 183--202, 2009.

\end{thebibliography}
\bibliographystyle{IEEEtran}

\end{document}